\documentclass[11pt,letterpaper]{article}
\usepackage[plain]{algorithm}
\usepackage{algorithmicx}
\usepackage[noend]{algpseudocode}
\usepackage{amssymb,amsmath}
\usepackage{latexsym}
\usepackage{listings}
\usepackage{times}

\title{Multinomial Loss on Held-out Data for the Sparse Non-negative Matrix Language Model}

\author{Ciprian Chelba and Fernando Pereira\\
  Google, Inc.\\
  1600 Amphitheatre Parkway\\
  Mountain View, CA 94043, USA\\
  {\tt \{ciprianchelba,pereira\}@google.com}}

\date{}

\begin{document}
\maketitle
\begin{abstract}
We describe Sparse Non-negative Matrix (SNM) language model estimation using multinomial loss on held-out data.

Being able to train on held-out data is important in practical situations where the training data
is usually mismatched from the held-out/test data. It is also less constrained than the previous 
training algorithm using leave-one-out on training data:
it allows the use of richer meta-features in the adjustment model, e.g.~the diversity counts used by Kneser-Ney
smoothing which would be difficult to deal with correctly in leave-one-out training.

In experiments on the one billion words language modeling benchmark~\cite{Chelba:2014}, we are able to slightly
improve on previous results reported in~\cite{Shazeer:2015}-\cite{Shazeer:2015a} which uses a different loss function, 
and employs leave-one-out training on a subset of the main training set. Surprisingly, an adjustment model with 
meta-features that discard all lexical information can perform as well as lexicalized meta-features. 
We find that fairly small amounts of held-out data (on the order of 30-70 thousand words) are sufficient for training the 
adjustment model.

In a real-life scenario where the training data is a mix of data sources that are imbalanced in size, and of different 
degrees of relevance to the held-out and test data, taking into account the data source for a given 
skip-/$n$-gram feature and combining them for best performance on held-out/test data improves over skip-/$n$-gram 
SNM models trained on pooled data by about 8\% in the SMT setup, or as much as 15\% in the ASR/IME setup.

The ability to mix various data sources based on how relevant they are to a mismatched held-out set is
probably the most attractive feature of the new estimation method for SNM LM.
\end{abstract}

\section{Introduction}
\label{sec:intro}

A statistical language model estimates probability values $P(W)$ for
strings of words $W$ in a vocabulary ${\mathcal V}$ whose size is in the
tens, hundreds of thousands and sometimes even millions.
Typically the string $W$ is broken into sentences, or other segments such
as utterances in automatic speech recognition, which are often assumed to be
conditionally independent; we will assume that $W$ is such a segment,
or sentence.

Since the parameter space of $P(w_k|w_1,w_2,\ldots,w_{k-1})$ is too
large, the language model is forced to put the \emph{context}\\
$W_{k-1}=w_1,w_2,\ldots,w_{k-1}$ into an \emph{equivalence class} determined
by a function $\Phi(W_{k-1})$. As a result,
\begin{equation}
\label{eq1}P(W)\cong\prod_{k=1}^nP(w_k|\Phi (W_{k-1}))
\end{equation}

Research in language modeling consists of finding appropriate
equivalence classifiers $\Phi$ and methods to estimate\\
$P(w_k|\Phi(W_{k-1}))$. Once the form $\Phi (W_{k-1})$ is specified, only the
problem of estimating $P(w_k|\Phi (W_{k-1}))$ from training data remains.

\subsection*{Perplexity as a Measure of Language Model Quality}

A \emph{statistical language model} can be evaluated by how well it
predicts a string of symbols $W_t$---commonly referred to as
\emph{test data}---generated by the source to be modeled.

A commonly used quality measure for a given model $M$ is related to
the entropy of the underlying source and was introduced under
the name of \emph{perplexity} (PPL):
\begin{eqnarray}
\label{basic_lm:ppl}
PPL(M) = exp\left(-\frac{1}{N} \sum_{k=1}^{N}\ln{[P_M(w_k|W_{k-1})]}\right)
\end{eqnarray}

For an excellent discussion on the use of perplexity in
statistical language modeling, as well as various estimates
for the entropy of English the reader is referred
to~\cite{jelinek97},~Section 8.4,~pages~141-142 and the additional reading
suggested in Section~8.5 of the same book.

\section{Notation and Modeling Assumptions}
\label{sec:notation}

We denote with $e$ an event in the training/development/test data corresponding to each prediction $(w_k|\Phi(W_{k-1}))$ in Eq.~(\ref{eq1}); each event consists of:
\begin{itemize}\addtolength{\itemsep}{-0.5\baselineskip}
\item a set of {\it features} $\mathcal{F}(e)=\{f_1, \ldots, f_k, \ldots, f_{F(e)}\} \subset \mathcal{F}$, where $\mathcal{F}$ denotes the set of features in the model, collected on the training data: $\mathcal{F} = \cup_{e \in \mathcal{T}} \mathcal{F}(e)$;
\item a {\it predicted (target) word} $w=t(e)$ from the LM vocabulary $\mathcal{V}$; we denote with $V=|\mathcal{V}|$ the size of the vocabulary.
\end{itemize}

The set of features $\mathcal{F}(e)$ is obtained by applying the equivalence classification function $\Phi(W_{k-1})$ to the context of the prediction. The most successful model so far has been the $n$-gram model, extracting all $n$-gram features of length $0, \ldots, n-1$ from the $W_{k-1}$ context\footnote{The empty feature is considered to have length 0, it is present in every event $e$, and it produces the unigram distribution on the language model vocabulary.}, respectively.

\subsection{Skip-$n$-gram Language Modeling}
\label{sec:skipgram}

A simple variant on the $n$-gram model is the skip-$n$-gram model; a skip-$n$-gram feature extracted from the context $W_{k-1}$ is characterized by the tuple $(r, s, a)$ where:
\begin{itemize}\addtolength{\itemsep}{-0.5\baselineskip}
\item $r$ denotes number of remote context words
\item $s$ denotes the number of skipped words
\item $a$ denotes the number of adjacent context words
\end{itemize}
relative to the target word $w_k$ being predicted.
For example, in the sentence,\\
\verb+<S> The quick brown fox jumps over the lazy dog </S>+\\
a $(1, 2, 3)$ skip-gram feature for the target word \verb+dog+ is:\\
\verb+[brown skip-2 over the lazy]+

To control the size of $\mathcal{F}(e)$ it is recommended to limit the skip length $s$ and also
either $(r + a)$ or both $r$ and $s$; not setting any such upper bounds
will result in events containing a set of skip-gram features whose total representation size is
quintic in the length of the sentence.

We configure the skip-$n$-gram feature extractor to produce all features $\mathbf{f}$, defined by
the equivalence class $\Phi(W_{k-1})$, that meet constraints on the minimum and maximum values for:
\begin{itemize}\addtolength{\itemsep}{-0.5\baselineskip}
\item the number of context words used $r + a$;
\item the number of remote words $r$;
\item the number of adjacent words $a$;
\item the skip length $s$.
\end{itemize}

We also allow the option of not including the exact value of $s$ in the
feature representation; this may help with smoothing by sharing counts
for various skip features. Tied skip-$n$-gram features will look like:\\
\verb+[curiosity skip-* the cat]+

Sample feature extraction configuration files for a $5$-gram and a skip-$10$-gram SNM LM are
presented in Appendix~\ref{appendix:5-gram-config}~and~\ref{appendix:skip-10-gram-config}, respectively.
A simple extension that leverages context beyond the current sentence,
as well as other categorical features such as geo-location is presented and evaluated
in~\cite{Chelba:ASRU2015}.

In order to build a good probability estimate for the target word $w_k$ in a context $W_{k-1}$,
or an event $e$ in our notation, we need a way of combining
an arbitrary number of features which do not fall into a simple
hierarchy like regular $n$-gram features. The following section describes
a simple, yet novel approach for combining such predictors in a way that
is computationally easy, scales up gracefully to large amounts of data
and as it turns out is also very effective from a modeling
point of view.

\section{Multinomial Loss for the Sparse Non-negative Matrix Language Model}
\label{sec:model}

The sparse non-negative matrix (SNM) language model (LM)~\cite{Shazeer:2015}-\cite{Shazeer:2015a}
assigns probability to a word by applying the equivalence classification function
$\Phi(W)$ to the context of the prediction, as explained in the previous section, and then using a
matrix $\textbf{M}$, where $M_{fw}$ is indexed by feature $f \in \mathcal{F}$ and
word $w \in \mathcal{V}$. We further assume that the model is parameterized as a slight variation
on conditional relative frequencies for words $w$ given features $f$, denoted as $c(w|f)$:
\begin{equation}
  P(w|\Phi(W)) \propto \sum_{f\in\Phi(W)} \underbrace{c(w|f) \cdot exp(A(f,w;\boldsymbol{\theta}))}_{M_{fw}}
\end{equation}
The \emph{adjustment function} $A(f,w;\boldsymbol{\theta})$ is a real-valued function whose task is to estimate the
relative importance of each input \emph{feature} $f$ for the prediction of the given \emph{target word} $w$.
It is computed by a linear model on \emph{meta-features} $h$ extracted from each \emph{link} $(f,w)$ and
associated feature $f$:
\begin{equation}
  A(f,w;\boldsymbol{\theta}) = \sum_k \theta_k h_k(f,w)\label{eq:adjustment}
\end{equation}
The meta-features are either strings identifying the feature type, feature, link etc., or bucketed feature and link counts. We also allow all possible conjunctions of elementary meta-features, and estimate a weight $\theta_k$ for each (elementary or conjoined) meta-feature $h_k$. In order to control the model size we use the hashing technique in~\cite{Ganchev:08},\cite{Weinberger:2009}. The meta-feature extraction is explained in more detail in Section~\ref{metafeature}, and associated Appendix~\ref{appendix}.

Assuming we have a sparse matrix $\textbf{M}$ of adjusted relative frequencies, the probability of
an event $e=(w|\Phi(W_{k-1}))$ predicting word $w$ in context $\Phi(W_{k-1})$ is computed as follows:
\begin{eqnarray}
P(e)   & = & y_t(e) / y(e) \nonumber\\
y_t(e) & = & \sum_{f \in \mathcal{F}} \sum_{w\in\mathcal{V}} 1_{f}(e) \cdot 1_{w}(e) M_{fw}\nonumber\\
y(e)   & = & \sum_{f \in \mathcal{F}} 1_{f}(e) M_{f*}\nonumber
\end{eqnarray}
where $M_{f*}$ ensures that the model is properly normalized over the LM vocabulary:
\begin{eqnarray}
M_{f*}  & = & \sum_{w\in\mathcal{V}} M_{fw}\nonumber
\end{eqnarray}
and the indicator functions $1_{f}(e)$ and $1_{w}(e)$ select a given feature, and target word in the event $e$, respectively:
\begin{eqnarray}
1_{f}(e) & = &
\begin{cases}
  1, f \in \mathcal{F}(e)\\
  0, o/w
\end{cases}\nonumber\\
1_{w}(e) & = &
\begin{cases}
  1, w = t(e)\\
  0, o/w
\end{cases}\nonumber
\end{eqnarray}

With this notation and using the shorthand $A_{fw} = A(f,w;\boldsymbol{\theta})$, the derivative of the log-probability for event $e$ with respect to the adjustment function $A_{fw}$ for a given link $(f,w)$ is:
\begin{eqnarray}
\frac{\partial \log P(e)}{\partial A_{fw}} & = & \frac{\partial \log y_t(e)}{\partial A_{fw}} - \frac{\partial \log y(e)}{\partial A_{fw}}\nonumber\\
& = & \frac{1}{y_t(e)}\frac{\partial y_t(e)}{\partial A_{fw}} - \frac{1}{y(e)}\frac{\partial y(e)}{\partial A_{fw}}\nonumber\\
& = & 1_{f}(e) M_{fw} \left[\frac{1_{w}(e)}{y_t(e)} - \frac{1}{y(e)}\right]\label{eq:grad}
\end{eqnarray}
making use of the fact that $\frac{\partial M_{fw}}{\partial A_{fw}} = \frac{\partial c(w|f) exp(A_{fw})}{\partial A_{fw}} = c(w|f) exp(A_{fw}) = M_{fw}$

Propagating the gradient $\frac{\partial \log P(e)}{\partial A_{fw}}$ to the $\boldsymbol{\theta}$ parameters of the adjustment function
$A(f,w;\boldsymbol{\theta})$ is done using mini-batch estimation for the reasons detailed in Section~\ref{sec:implementation}:
\begin{eqnarray}
  \frac{\partial \log P(e)}{\partial \theta_k} & = & \sum_{(f,w): h_k \in meta-features(f,w)} \frac{\partial \log P(e)}{\partial A_{fw}}\nonumber\\
  \theta_{k,B+1} & \leftarrow & \theta_{k,B} - \eta \sum_{e \in B} \frac{\partial \log P(e)}{\partial \theta_k} \label{eq:thetagrad}
\end{eqnarray}

Rather than using a single fixed learning rate $\eta$, we use AdaGrad~\cite{Duchi:2011} which uses a separate adaptive learning rate $\eta_{k,B}$ for each weight $\theta_{k,B}$:
\begin{equation}
  \eta_{k,B} = \frac{\gamma}{\sqrt{\Delta_0 + \sum_{b=1}^B \left[\sum_{e \in b} \frac{\partial \log P(e)}{\partial \theta_k}\right]^2}}\label{eq:adagrad}
\end{equation}
where $B$ is the current batch index, $\gamma$ is a constant scaling factor for all learning rates and $\Delta_0$ is an initial accumulator constant. Basing the learning rate on historical information tempers the effect of
frequently occurring features which keeps the weights small and as such acts as a form of regularization.

\subsection{Implementation Notes}
\label{sec:implementation}

From a computational point of view, the two main issues with a straightforward gradient descent parameter update
(either on-line or batch) are:
\begin{enumerate}\addtolength{\itemsep}{-0.5\baselineskip}
\item the second term on the right-hand-side (RHS) of Eq.~(\ref{eq:grad}) is an update that needs to be propagated
to all words in the vocabulary, irrespective of whether they occur on a given training event or not;
\item keeping the model normalized after a $M_{fw}$ parameter update means recomputing all normalization
coefficients $M_{f*}, \forall f \in \mathcal{F}$.
\end{enumerate}
For mini-/batch updates, the model renormalization is done at the end of each training epoch/iteration, and it is
no longer a problem. To side-step the first issue, we notice that mini-/batch updates would allow us to accumulate the
$\alpha_f(B) = \sum_{e \in B} 1_{f}(e) \frac{1}{y(e)}$ across the entire mini-/batch $B$, and adjust
the cumulative gradient at the end of the mini-/batch, in effect computing:
\begin{eqnarray}
\sum_{e \in B} \frac{\partial \log P(e)}{\partial A_{fw}} & = & \sum_{e \in B} M_{fw} \frac{1}{y_t(e)} 1_{f}(e) \cdot 1_{w}(e) - M_{fw} \cdot \alpha_f(B)\label{eq:batch_update}\\
\alpha_f(B) & = & \sum_{e \in B} 1_{f}(e) \frac{1}{y(e)}\nonumber
\end{eqnarray}

In summary, we use two maps to compute the gradient updates over a mini-/batch: one keyed by $(f,w)$ pairs, and one keyed by $f$. The first map accumulates the first term on the RHS of Eq.~(\ref{eq:batch_update}), and is updated once for each link $(f,w)$ occurring in a training event $e$. The second map accumulates the $\alpha_f(B)$ values, and is again updated only for each of the features $f$ encountered on a given event in the mini-/batch. At the end of the mini-/batch we update the entries in the first map acc.\ to Eq.~(\ref{eq:batch_update}) such that they store the cumulative gradient; these are then used to update the $\boldsymbol{\theta}$ parameters of the adjustment function according to Eq.~(\ref{eq:thetagrad}).

The model $M_{fw}$ and the normalization coefficients $M_{f*}$ are stored in maps keyed by $(f,w)$, and $f$, respectively. The $\left[(f,w) \Rightarrow M_{fw}\right]$ map is initialized with relative frequencies $c(w|f)$ computed from the training data; on disk they are stored in an SSTable~\cite{Chang:2008} keyed by $(f,w)$, with $f$ and $w$ represented as plain strings. For training the adjustment model we only need the rows of the $\textbf{M}$ matrix that are encountered on development data (i.e., the training data for the adjustment model). A MapReduce~\cite{Ghemawat:2004} with two inputs extracts and intersects the features encountered on development data with the features collected on the main training data---where the relative frequencies $c(w|f)$ were also computed. The output is a significantly smaller matrix $\textbf{M}$ that is loaded in RAM and used to train the adjustment model.

\subsection{Meta-features extraction}
\label{metafeature}
The process of breaking down the original features into meta-features and recombining them, allows similar features, i.e.\ features that are different only in some of their base components, to share weights, thus improving generalization. 

Given an event \texttt{the quick brown fox}, the 4-gram feature for the prediction of the target \texttt{fox} would be broken down into the following elementary meta-features:
\begin{itemize}\addtolength{\itemsep}{-0.5\baselineskip}
\item feature identity, e.g.\ \texttt{[the quick brown]}
\item feature type, e.g.\ 3-gram
\item feature count $C_{f*}$
\item target identity, e.g.\ \texttt{fox}
\item feature-target count $C_{fw}$
\end{itemize}

Elementary meta-features of different types are then joined with others to form more complex meta-features, as described best by the pseudo-code in Appendix~\ref{appendix}; note that the seemingly
absent feature-target identity is represented by the conjunction of the feature identity and the target identity.

As count meta-features of the same order of magnitude carry similar information, we group them so they can share weights.
We do this by bucketing the count meta-features according to their (floored) $\log_2$ value. Since this effectively puts
the lowest count values, of which there are many, into a different bucket, we optionally introduce a second (ceiled)
bucket to assure smoother transitions. Both buckets are then weighted according to the $\log_2$ fraction lost by
the corresponding rounding operation.

To control memory usage, we employ a feature hashing
technique~\cite{Ganchev:08},\cite{Weinberger:2009} where we store the meta-feature weights in a flat hash table of 
predefined size; strings are fingerprinted, counts are hashed and the resulting integer mapped to an index $k$ in 
$\boldsymbol{\theta}$ by taking its value modulo the pre-defined $size(\boldsymbol{\theta})$. We do not prevent collisions, 
which has the potentially undesirable effect of tying together the weights of different meta-features. 
However, when this happens the most frequent meta-feature will dominate the final value after training, which essentially 
boils down to a form of pruning. Because of this the model performance does not strongly depend on the size of the hash table.

\section{Experiments}

\subsection{Experiments on the One Billion Words Language Modeling Benchmark}

Our first experimental setup used the One Billion Word Benchmark corpus\footnote{http://www.statmt.org/lm-benchmark}
made available by~\cite{Chelba:2014}.
For completeness, here is a short description of the corpus, containing only monolingual English data:
\begin{itemize}\addtolength{\itemsep}{-0.5\baselineskip}
\item Total number of training tokens is about 0.8 billion
\item The vocabulary provided consists of 793471 words including sentence boundary markers
\texttt{<S>}, \texttt{</S>}, and was constructed by discarding all words with count below 3
\item Words outside of the vocabulary were mapped to an \texttt{<UNK>} token, also part of the vocabulary
\item Sentence order was randomized
\item The test data consisted of 159658 words (without counting the sentence beginning marker
\texttt{<S>} which is never predicted by the language model)
\item The out-of-vocabulary (OOV) rate on the test set was 0.28\%.
\end{itemize}

The foremost concern when using held-out data for estimating the adjustment model is the limited amount of
data available in a practical setup, so we used a small development set consisting of 33 thousand words.

We conducted experiments using two feature extraction configurations identical to those used in~\cite{Shazeer:2015}:\\
\texttt{5-gram} and \texttt{skip-10-gram}, see Appendix~\ref{appendix:5-gram-config}~and~\ref{appendix:skip-10-gram-config}.
The AdaGrad parameters in Eq.~(\ref{eq:adagrad}) are set to:
$\gamma=0.1$, $\Delta_0 = 1.0$, and the mini-batch size is 2048 samples. We also experimented with various
adjustment model sizes (200M, 20M, and 200k hashed parameters), non-lexicalized meta-features, and feature-only meta-features,
see Appendix~\ref{appendix}. The results are presented in Tables~\ref{tab:5-gram-exps}-\ref{tab:skip-10-gram-exps}.

A first conclusion is that we can indeed get away with very small amounts of development data. This is excellent news,
because usually people do not have lots of development data to tune parameters on, see SMT experiments presented in the
next section. Using meta-features computed only from the feature component of a link does lead to a fairly significant
increase in PPL: 5\% rel for the \texttt{5-gram} config, and 10\% rel for the \texttt{skip-10-gram} config.

Surprisingly, when using the \texttt{5-gram} config, discarding the lexicalized meta-features consistently
does a tiny bit better than the lexicalized model; for the \texttt{skip-10-gram} config
the un-lexicalized model performs essentially as well as the lexicalized model. The number of parameters in the
model is very small in this case (on the order of a thousand) so the model no longer over-trains after the first iteration
as was the case when using link lexicalized meta-features; meta-feature hashing is not necessary either.

In summary, training and evaluating in exactly the same training/test setup as the one in~\cite{Shazeer:2015} we find that:
\begin{enumerate}\addtolength{\itemsep}{-0.5\baselineskip}
\item \texttt{5-gram} config: using multinomial loss training on 33 thousand words of development data, 200K or larger adjustment model, and un-lexicalized
meta-features trained over 5 epochs produces \texttt{5-gram} SNM PPL of 69.6, which is just a bit better than the
\texttt{5-gram} SNM PPL of 70.8 reported in~\cite{Shazeer:2015}, Table~1, and very close to the Kneser-Ney PPL of 67.6.
\item \texttt{skip-10-gram} config: using multinomial loss training on 33 thousand words of development data, 20M or larger adjustment model, and un-lexicalized
meta-features trained over 5 epochs produced \texttt{skip-10-gram} SNM PPL of 50.9, again just a bit better than both the \texttt{skip-10-gram}
SNM PPL of 52.9 and the RNN-LM PPL of 51.3 reported in~\cite{Shazeer:2015}, Table~3, respectively.
\end{enumerate}

\begin{table}
\centering
\begin{tabular}{|l|l|l|l|r|r|}
\hline
Model Size  & Num Training & \multicolumn{2}{c}{Metafeatures Extraction}& Test Set & Actual Num Hashed Params   \\
(max num hashed params) & Epochs       & lexicalized & feature-only                 & PPL      & (non-zero) \\\hline
0           & \multicolumn{3}{l|}{\emph{Unadjusted Model}}                     & 86.0     &           0        \\\hline
200M        & 1            & yes         & no                           & 71.4     &   116205951        \\
            &              & yes         & yes			        & 75.8     &       72447        \\
            &              & no          & no                           & 70.3     &         567        \\\cline{2-6}
            & 5            & yes         & no                           & 78.7     &   116205951        \\
            &              & yes         & yes			        & 73.9     &       72447        \\
            &              & no          & no                           & \bf{69.6}&         567        \\\hline
20M         & 1            & yes         & no                           & 71.4     &    20964888        \\
            &              & yes         & yes			        & 75.8     &       72344        \\
            &              & no          & no                           & 70.3     &         567        \\\cline{2-6}
            & 5            & yes         & no                           & 78.8     &    20964888        \\
            &              & yes         & yes			        & 73.9     &       72447        \\
            &              & no          & no                           & \bf{69.6}&         567        \\\hline
200K        & 1            & yes         & no                           & 72.0     &      204800        \\
            &              & yes         & yes			        & 75.9     &       61022        \\
            &              & no          & no                           & 70.3     &         566        \\\cline{2-6}
            & 5            & yes         & no                           & 84.8     &      204800        \\
            &              & yes         & yes			        & 73.9     &       61022        \\
            &              & no          & no                           & \bf{69.6}&         567        \\\hline
\end{tabular}
\caption{Experiments on the One Billion Words Language Modeling Benchmark in \texttt{5-gram} configuration; 2048 mini-batch size, one and five training epochs.}
\label{tab:5-gram-exps}
\end{table}

\begin{table}
\centering
\begin{tabular}{|l|l|l|l|r|r|}
\hline
Model Size  & Num Training & \multicolumn{2}{c}{Metafeatures Extraction}& Test Set & Actual Num Hashed Params   \\
(max num hashed params) & Epochs       & lexicalized & feature-only                 & PPL      & (non-zero) \\\hline
0           & \multicolumn{3}{l|}{\emph{Unadjusted Model}}                     & 69.2     &           0        \\\hline
200M        & 1            & yes         & no                           & 52.2     &   209234366        \\
            &              & yes         & yes			        & 58.0     &      740836        \\
            &              & no          & no                           & 52.2     &        1118        \\\cline{2-6}
            & 5            & yes         & no                           & 54.3     &   209234366        \\
            &              & yes         & yes			        & 56.1     &      740836        \\
            &              & no          & no                           & \bf{50.9}&        1118        \\\hline
20M         & 1            & yes         & no                           & 52.2     &    20971520        \\
            &              & yes         & yes			        & 58.0     &      560006        \\
            &              & no          & no                           & 52.2     &        1117        \\\cline{2-6}
            & 5            & yes         & no                           & 54.4     &    20971520        \\
            &              & yes         & yes			        & 56.1     &      560006        \\
            &              & no          & no                           & \bf{50.9}&        1117        \\\hline
200K        & 1            & yes         & no                           & 52.4     &      204800        \\
            &              & yes         & yes			        & 58.0     &      194524        \\
            &              & no          & no                           & 52.2     &        1112        \\\cline{2-6}
            & 5            & yes         & no                           & 56.5     &      204800        \\
            &              & yes         & yes			        & 56.1     &      194524        \\
            &              & no          & no                           & 51.0     &        1112        \\\hline
\end{tabular}
\caption{Experiments on the One Billion Words Language Modeling Benchmark in \texttt{skip-10-gram} configuration; 
2048 mini-batch size, one and five training epochs.}
\label{tab:skip-10-gram-exps}
\end{table}

\subsection{Experiments on 10B Words of Burmese Data in Statistical Machine Translation Language Modeling Setup}
In a separate set of experiments on Burmese data provided by the statistical machine translation (SMT) team, 
the held-out data (66 thousand words) and the test data (22 thousand words) is mismatched to the training data 
consisting of 11 billion words mostly crawled from the web (and labelled as ``web'') along with 176 million words 
(labelled as ``target'') originating from parallel data used for training the channel model. The vocabulary size
is 785261 words including sentence boundary markers; the out-of-vocabulary rate on both held-out and test set is 0.6\%.

To quantify statistically the mismatch between training and held-out/test data, we trained both Katz and interpolated Kneser-Ney \texttt{5-gram} 
models on the pooled training data; the Kneser-Ney LM has PPL of 611 and 615 on the held-out and test data, respectively; 
the Katz LM is severely more mismatched, with PPL of 4153 and 4132, respectively\footnote{The cummulative hit-ratios on test data at orders 
5 through 1 were 0.2/0.3/0.6/0.9/1.0 for the KN model, and 0.1/0.3/0.6/0.9/1.0 for the Katz model, which may explain the large gap 
in performance between KN and Katz: the diversity counts used by KN 80\% of the time are more robust to mismatched training/test conditions 
than the relative frequencies used by Katz.}.
Because of the mismatch between the training and the held-out/test data, the PPL of the un-adjusted SNM \texttt{5-gram} LM is
significantly lower than that of the SNM adjusted using leave-one-out~\cite{Shazeer:2015} on a subset of the shuffled
training set: 710 versus 1285.

The full set of results in this experimental setup are presented in Tables~\ref{tab:smt-snm-5-gram-exps}-\ref{tab:smt-snm-skip-5-gram-exps}.

When using the multinomial adjustment model training on held-out data things fall in place, and the adjusted
SNM \texttt{5-gram} has lower PPL than the unadjusted one: 347 vs 710; the former is significantly lower
than the 1285 value produced by leave-one-out training; the \texttt{skip-5-gram} SNM model (a trimmed down 
version of the \texttt{skip-10-gram} in Appendix~\ref{appendix:skip-10-gram-config}) has PPL of 328, 
improving only modestly over the 5-gram SNM result---perhaps due to the mismatch between training and development/test data.

We also note that the lexicalized adjustment model works significantly better then either the feature-only or the un-lexicalized one, 
in contrast to the behavior on the one billion words benchmark.

As an extension we experimented with SNM training that takes into account the data source for a given
skip-/$n$-gram feature, and combines them best on held-out/test data by taking into account the identity of
the data source as well. This is the reality of most practical scenarios for training language models.
We refer to such features as \texttt{corpus-tagged} features: in training we augment each feature with a tag describing the 
training corpus it originates from, in this case \texttt{web} and \texttt{target}, respectively; on held-out and test data 
the event extractor augments each feature with each of the corpus tags in training. The adjustment function is then trained 
to assign a weight for each such \texttt{corpus-tagged} feature.
Corpus tagging the features and letting the adjustment model do the combination reduced PPL by about 8\% relative 
over the model trained on pooled data in both \texttt{5-gram} and \texttt{skip-5-gram} configurations.

\subsection{Experiments on 35B Words of Italian Data in Language Modeling Setup for Automatic Speech Recognition}

We have experimented with SNM LMs in the LM training setup for Italian as used on the automatic speech recognition (ASR) project. 
The same LM is used for two distinct types of ASR requests: voice-search queries (VS) and Android Input Method (IME, speech 
input on the soft keyboard). As a result we use two separate test sets to evaluate the LM performance, one for each VS and IME,
respectively. 

The held-out data used for training the adjustment model is a mix of VS and IME transcribed utterances, consisting of 
36 thousand words split 30/70\% between VS/IME, respectively. The adjustment model used 20 million parameters trained using
mini-batch AdaGrad (2048 samples batch size) in one epoch.

The training data consists of a total of 35 billion words from various sources, of varying size and degree of relevance
to either of the test sets:
\begin{itemize}
\item \verb+google.com+ (111 Gbytes) and \verb+maps.google.com+ (48 Gbytes) query stream
\item high quality web crawl (5 Gbytes)
\item automatically transcribed utterances filtered by ASR confidence for both VS and IME (4.1 and 0.5 Gbytes, respectively)
\item manually transcribed utterances for both VS and IME (0.3 and 0.5 Gbytes, repectively)
\item voice actions training data (0.1 Gbytes)
\end{itemize}

As a baseline for the SNM we built Katz and interpolated Kneser-Ney 5-gram models by pooling all the training data. We then
built a 5-gram SNM LM, as well as corpus-tagged SNM 5-gram where each $n$-gram is tagged with the identity of the corpus
it occurs in (one of seven tags). Skip-grams were added to either of the SNM models. The results are presented in
Table~\ref{tab:snm_it}; the vocabulary used to train all language models being compared consisted of 4 million words.

\begin{table}
\centering
\begin{tabular}{|l|r|r|}
\hline
Model & \multicolumn{2}{r|}{Perplexity}\\\cline{2-3}
      & IME & VS\\\hline
Katz 5-gram                                                                                         & 177 & 154\\
Interpolated Kneser-Ney 5-gram                                                                      & 152 & 142\\\hline
SNM 5-gram, adjusted                                                                                & 104 & 126\\
SNM 5-gram, corpus-tagged, adjusted                                                                 &  88 & 124\\
SNM 5-gram, skip-gram, adjusted                                                                     &  96 & 119\\
SNM 5-gram, skip-gram, corpus-tagged, adjusted                                                      &  86 & 119\\\hline
\end{tabular}
\caption{Perplexity Results of Various Approaches to Language Modeling in the Setup Used for Italian ASR.}
\label{tab:snm_it}
\end{table}

A first observation is that the SNM 5-gram LM outperforms both Katz and Kneser-Ney LMs significantly on both test sets. 
We attribute this to the ability of the adjustment model to optimize the combination of various $n$-gram contexts such 
that they maximize the likelihood of the held-out data; no such information is available to either of the Katz/Kneser-Ney models.

Augmenting the SNM 5-gram with corpus-tags benefits mostly the IME performance; we attribute this to the fact that the vast
majority of the training data is closer to the VS test set, and clearly separating the training sources (in particular the ones
meant for the IME component of the LM such as web crawl and IME transcriptions) allows the adjustment model to optimize 
better for that subset of the held-out data. Skip-grams offer relatively modest improvements over either SNM 5-gram models.

\section{Conclusions and Future Work}
The main conclusion is that training the adjustment model on held-out data using multinomial loss introduces many advantages while matching
the previous results reported in~\cite{Shazeer:2015}: as observed in~\cite{Xu:2011}, Section 2, using a binary probability
model is expected to yield the same model as a multinomial probability model. Correcting the deficiency in~\cite{Shazeer:2015}
induced by using a Poisson model for each binary random variable does not seem to make a difference in the quality
of the estimated model.

Being able to train on held-out data is very important in practical situations where the training data is
usually mismatched from the held-out/test data.
It is also less constrained than the previous training algorithm using leave-one-out on training data:
it allows the use of richer meta-features in the adjustment model, e.g.~the diversity counts used by Kneser-Ney
smoothing which would be difficult to deal with correctly in leave-one-out training. Taking into account the data source for a given 
skip-/$n$-gram feature, and combining them for best performance on held-out/test data improves over SNM models trained on pooled data by about 8\% in the SMT setup, or as much as 15\% in the ASR/IME setup.

We find that fairly small amounts of held-out data (on the order of 30-70 thousand words) are sufficient for training the 
adjustment model. Surprisingly, using meta-features that discard all lexical information can sometimes perform as well 
as lexicalized meta-features, as demonstrated by the results on the One Billion Words Benchmark corpus.

Given the properties of the SNM $n$-gram LM explored so far:
\begin{itemize}
\item ability to mix various data sources based on how relevant they are to a given held-out set, thus providing an alternative to 
Bayesian mixing algorithms such as~\cite{Allauzen:2011},
\item excellent pruning properties relative to entropy pruning of Katz and Kneser-Ney models~\cite{Pelemans:2015},
\item conversion to standard ARPA back-off format~\cite{Pelemans:2015},
\item effortless incorporation of richer features such as skip-$n$-grams and geo-tags~\cite{Chelba:ASRU2015},
\end{itemize}
we believe SNM could provide the estimation back-bone for a fully fledged LM training pipeline used in a real-life setup.

A comparison of SNM against maximum entropy modeling at feature extraction parity is also long due.

\newpage
\begin{table}[h!]
\centering
\begin{tabular}{|l|l|l|l|r|r|}
\hline
Model Size  & Num Training & \multicolumn{2}{c}{Metafeatures Extraction}& Test Set & Actual Num Hashed Params   \\
(max num hashed params) & Epochs       & lexicalized & feature-only                 & PPL      & (non-zero) \\\hline\hline
\multicolumn{2}{|l}{\bf{Leave-one-out}} &     &                         &      &                    \\\cline{1-1}
200M        & ---          & yes         & no                           & 1285 &                    \\\hline\hline
\multicolumn{2}{|l}{\bf{Multinomial}}    &     &                        &      &                    \\\cline{1-1}
0           & \multicolumn{3}{l|}{\emph{Unadjusted Model}}                     & 710  &           0        \\\hline
200M        & 1            & yes         & no                           &  352 &            103549851   \\
            &              & yes         & yes			        &  653 &            87875   \\
            &              & no          & no                           &  569 &            716     \\\cline{2-6}
            & 5            & yes         & no                           &  \underline{347} & 103549851\\
            &              & yes         & yes			        &  638 &             87875\\
            &              & no          & no                           &  559 &             716\\\hline
20M         & 1            & yes         & no                           &  353 &            20963883 \\
            &              & yes         & yes			        &  653 &            87712   \\
            &              & no          & no                           &  569 &            716     \\\cline{2-6}
            & 5            & yes         & no                           &  348 &            20963883\\
            &              & yes         & yes			        &  638 &            87712   \\
            &              & no          & no                           &  559 &            716     \\\hline
200K        & 1            & yes         & no                           &  371 &            204800  \\
            &              & yes         & yes			        &  653 &            71475   \\
            &              & no          & no                           &  569 &            713     \\\cline{2-6}
            & 5            & yes         & no                           &  400 &            204800  \\
            &              & yes         & yes			        &  638 &            71475   \\
            &              & no          & no                           &  560 &            713     \\\hline\hline
\multicolumn{2}{|l}{\bf{Multinomial, corpus-tagged}}    &     &         &      &                    \\\cline{1-1}
0           & \multicolumn{3}{l|}{\emph{Unadjusted Model}}                     & 574  &            0        \\\hline
200M        & 1            & yes         & no                           & \bf{323}  &       129291753\\
            &              & yes         & yes			        & 502  &            157684\\
            &              & no          & no                           & 447  &            718\\\cline{2-6}
            & 5            & yes         & no                           & 324  &            129291753\\
            &              & yes         & yes			        & 488  &            157684\\
            &              & no          & no                           & 442  &            718\\\hline
20M         & 1            & yes         & no                           & \bf{323}  &       20970091\\
            &              & yes         & yes			        & 502  &            157141\\
            &              & no          & no                           & 447  &            718\\\cline{2-6}
            & 5            & yes         & no                           & 324  &            20970091\\
            &              & yes         & yes			        & 488  &            157141\\
            &              & no          & no                           & 442  &            718\\\hline
200K        & 1            & yes         & no                           & 334  &            204800\\
            &              & yes         & yes			        & 502  &            110150\\
            &              & no          & no                           & 447  &            715\\\cline{2-6}
            & 5            & yes         & no                           & 356  &            204800\\
            &              & yes         & yes			        & 489  &            110150\\
            &              & no          & no                           & 442  &            715\\\hline
\end{tabular}
\caption{SMT Burmese Dataset experiments in \texttt{5-gram} configuration, with and without \texttt{corpus-tagged} feature extraction; 2048 mini-batch size, one and five training epochs.}
\label{tab:smt-snm-5-gram-exps}
\end{table}

\begin{table}
\centering
\begin{tabular}{|l|l|l|l|r|r|}
\hline
Model Size  & Num Training & \multicolumn{2}{c}{Metafeatures Extraction}& Test Set & Actual Num Hashed Params   \\
(max num hashed params) & Epochs       & lexicalized & feature-only                 & PPL      & (non-zero) \\\hline\hline
\multicolumn{2}{|l}{\bf{Multinomial}}    &     &                        &      &                    \\\cline{1-1}
0           & \multicolumn{3}{l|}{\emph{Unadjusted Model}}                     & 687  &           0        \\\hline
200M        & 1            & yes         & no                           & \underline{328}  &  209574343\\
            &              & yes         & yes			        & 587  &  772743\\
            &              & no          & no                           & 496  &  1414\\\hline
20M         & 1            & yes         & no                           & 328  &  20971520\\
            &              & yes         & yes			        & 587  &  760066\\
            &              & no          & no                           & 496  &  1414\\\hline
200K        & 1            & yes         & no                           & 342  &  204800\\
            &              & yes         & yes			        & 587  &  200060\\
            &              & no          & no                           & 496  &  1408\\\hline\hline
\multicolumn{2}{|l}{\bf{Multinomial, corpus-tagged}}    &     &         &      &                    \\\cline{1-1}
0           & \multicolumn{3}{l|}{\emph{Unadjusted Model}}                     & 567  &           0        \\\hline
200M        & 1            & yes         & no                           & \bf{302}  &  209682449\\
            &              & yes         & yes			        & 474  &  1366944\\
            &              & no          & no                           & 405  &  1416\\\hline
20M         & 1            & yes         & no                           & 303  &  20971520\\
            &              & yes         & yes			        & 474  &  1327393\\
            &              & no          & no                           & 405  &  1416\\\hline
200K        & 1            & yes         & no                           & 312  &  204800\\
            &              & yes         & yes			        & 474  &  204537\\
            &              & no          & no                           & 405  &  1409\\\hline
\end{tabular}
\caption{SMT Burmese Dataset experiments in \texttt{skip-5-gram} configuration, with and without \texttt{corpus-tagged} feature extraction; 
2048 mini-batch size, one training epoch.}
\label{tab:smt-snm-skip-5-gram-exps}
\end{table}

\section{Acknowledgments}
Thanks go 
to Yoram Singer for clarifying the correct mini-batch variant of AdaGrad, Noam Shazeer for assistance on
understanding his implementation of the adjustment function estimation, Diamantino Caseiro for code reviews,
Kunal Talwar, Amir Globerson and Diamantino Caseiro for useful discussions, and Anton Andryeyev for providing
the SMT training/held-out/test data sets. Last, but not least, we are thankful to our former summer intern 
Joris Pelemans for suggestions while preparing the final version of the paper.

\appendix
\section{Appendix: \texttt{5-gram} Feature Extraction Configuration}
\label{appendix:5-gram-config}
\begin{lstlisting}[language=C++]
// Sample config generating a straight 5-gram language model.
ngram_extractor {
  min_n: 0
  max_n: 4
}
\end{lstlisting}

\section{Appendix: \texttt{skip-10-gram} Feature Extraction Configuration}
\label{appendix:skip-10-gram-config}
\begin{lstlisting}[language=C++]
// Sample config generating a straight skip-10-gram language model.
ngram_extractor {
  min_n: 0
  max_n: 9
}
skip_ngram_extractor {
  max_context_words: 4
  min_remote_words: 1
  max_remote_words: 1
  min_skip_length: 1
  max_skip_length: 10
  tie_skip_length: true
}
skip_ngram_extractor {
  max_context_words: 5
  min_skip_length: 1
  max_skip_length: 1
  tie_skip_length: false
}\end{lstlisting}

\newpage
\section{Appendix: Meta-features Extraction Pseudo-code}
\label{appendix}
\begin{algorithm}
\begin{algorithmic}
\State{// Metafeatures are represented as tuples (hash\_value, weight).}
\State{// Concat(metafeatures, end\_pos, mf\_new) concatenates mf\_new}
\State{// with all the existing metafeatures up to end\_pos.}
\Function{ComputeMetafeatures}{FeatureTargetPair pair}
    \State{// feature-related metafeatures}
    \State{metafeatures $\gets$ (Fingerprint(pair.feature.id()), 1.0)}
    \State{metafeatures $\gets$ (Fingerprint(pair.feature.type()), 1.0)}
    \State{ln\_count = log(pair.feature.count()) / log(2)}
    \State{bucket1 = floor(ln\_count)}
    \State{bucket2 = ceil(ln\_count)}
    \State{weight1 = bucket2 - ln\_count}
    \State{weight2 = ln\_count - bucket1}
    \State{metafeatures $\gets$ (Hash(bucket1), weight1)}
    \State{metafeatures $\gets$ (Hash(bucket2), weight2)}
    \State{}
    \State{// target-related metafeatures}
    \State{Concat(metafeatures, metafeatures.size(), (Fingerprint(pair.target.id()), 1.0))}
    \State{}
    \State{// feature-target-related metafeatures}    
    \State{ln\_count = log(pair.count()) / log(2)}
    \State{bucket1 = floor(ln\_count)}
    \State{bucket2 = ceil(ln\_count)}
    \State{weight1 = bucket2 - ln\_count}
    \State{weight2 = ln\_count - bucket1}
    \State{Concat(metafeatures, metafeatures.size(), (Hash(bucket1), weight1))}
    \State{Concat(metafeatures, metafeatures.size(), (Hash(bucket2), weight2))}
    \State{}    
    \State{return metafeatures}
\EndFunction
\end{algorithmic}
\end{algorithm}

\end{document}